\documentclass[11pt]{article}

\usepackage[final]{acl}

\usepackage{times}
\usepackage{latexsym}
\usepackage{CJKutf8}
\usepackage[T1]{fontenc}
\usepackage[utf8]{inputenc}

\usepackage{microtype}
\usepackage{inconsolata}

\usepackage{graphicx}
\usepackage{amsmath}
\usepackage{amssymb}
\usepackage{amsfonts}
\usepackage{booktabs}
\usepackage{multirow}
\usepackage{enumitem}
\usepackage{algorithm}
\usepackage{algorithmic}
\usepackage{xcolor}
\usepackage{url}
\usepackage{arydshln}  

\title{\textbf{CSRP: Chain-of-Thought Reasoning for Chinese Text Correction via Reinforcement Learning with Efficiency-Aware Rewards}}
\author{
  Wei Tian\textsuperscript{1}, 
  Yuhao Zhou\textsuperscript{1}, 
  Man Lan\textsuperscript{1,2}\thanks{\ \ Corresponding author.} \\
  \textsuperscript{1}School of Computer Science and Technology, East China Normal University \\
  \textsuperscript{2}Shanghai Institute of Artificial Intelligence for Education, East China Normal University \\
  \texttt{\{tianwei, yhzhou\}@stu.ecnu.edu.cn, mlan@cs.ecnu.edu.cn}
}

\date{}

\begin{document}
\maketitle

\begin{abstract}
Large Language Model (LLM) based Chinese Grammatical Error Correction (CGEC) systems face two critical challenges: general-purpose models lack specialized linguistic priors for subtle grammatical distinctions, and Supervised Fine-Tuning (SFT) with Maximum Likelihood Estimation fails to optimize for precision-focused metrics, leading to systematic over-correction. We propose CSRP, a three-stage framework that progressively builds correction capability through Continual Pre-training (CPT) on 5.9M balanced samples to internalize domain knowledge, Chain-of-Thought SFT with explicit error reasoning for diagnostic transparency, and Group Relative Policy Optimization with a novel Efficiency-Aware Reward that explicitly penalizes unnecessary edits. On the NACGEC benchmark, CSRP achieves state-of-the-art performance with 50.99 $F_{0.5}$ and 57.17 precision, substantially outperforming previous best results while effectively mitigating the over-correction bias inherent in MLE-trained models. Our method also advances CSCD spelling correction to 59.61 F1, surpassing GPT-4 by 5.20 points. Comprehensive ablation studies demonstrate that the RL alignment stage contributes a 8\% relative gain over the SFT baseline, and that this gain is orthogonal to the contribution of large-scale CPT, validating that explicit optimization for edit efficiency is essential for high-quality grammatical error correction. Our code is available at \url{https://github.com/TW-NLP/ChineseErrorCorrector}.
\end{abstract}

\section{Introduction}
\begin{figure}[!t]
    \centering
    \includegraphics[width=1.0\linewidth]{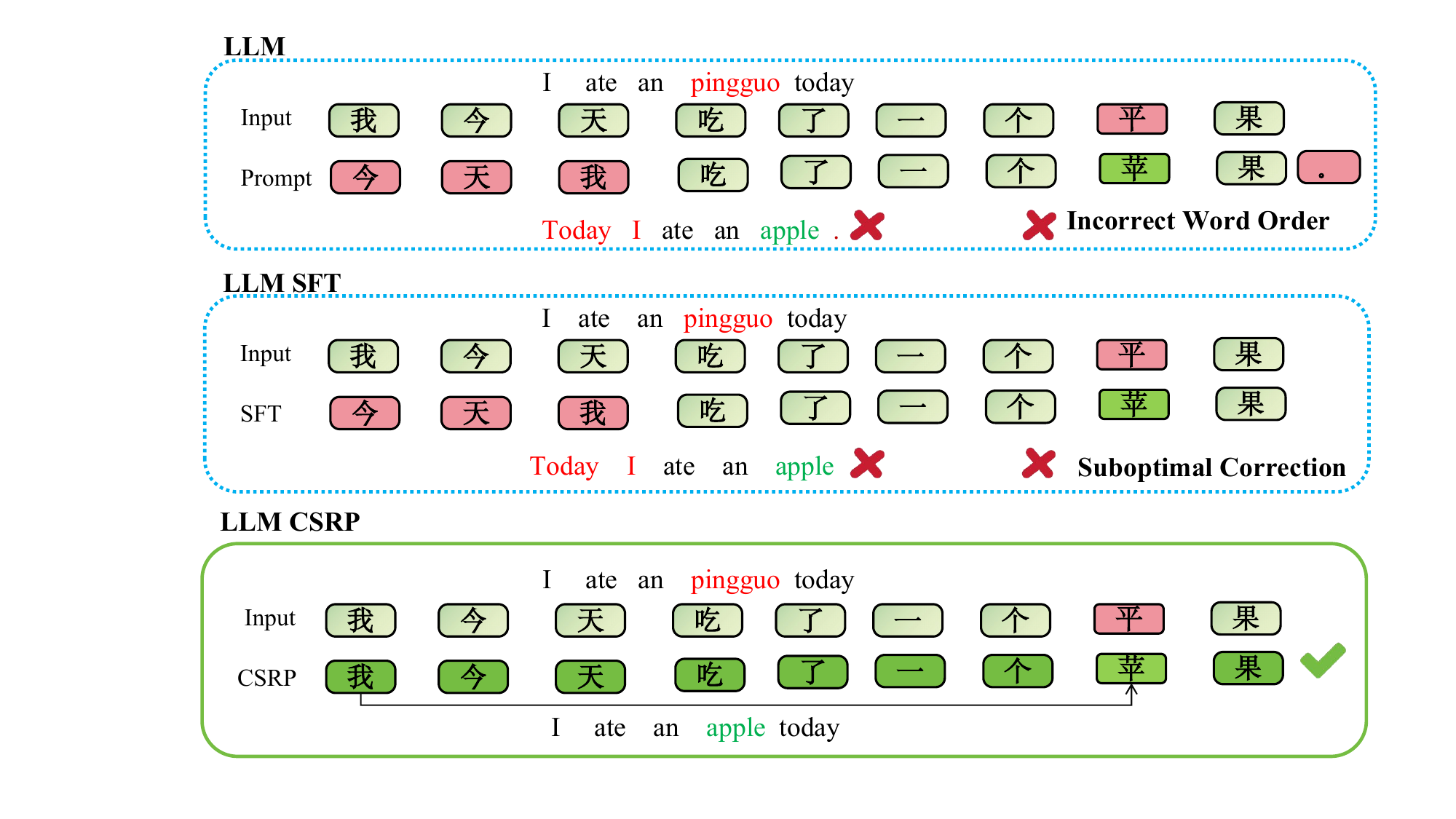}
    \caption{Supervised Fine-Tuning performance plateau.}
    \label{fig:example_problem}
\end{figure}
The essence of Grammatical Error Correction (GEC) lies in repairing linguistic deviations while strictly preserving the semantic fidelity of the original text. An ideal GEC system should exhibit high \textbf{faithfulness}, adhering to the principle of \textbf{minimal editing} by intervening only when necessary. However, despite the potential demonstrated by Large Language Models (LLMs) in Chinese GEC (CGEC) through their generative capabilities, the prevailing paradigm based on Supervised Fine-Tuning (SFT) has encountered a significant \textbf{performance plateau}, as shown in Figure~\ref{fig:example_problem}. Mainstream models consistently stagnate within an $F_{0.5}$ score range of 45-46 on authoritative benchmarks\cite{ChineseErrorCorrector3}.

Our investigation reveals that this limitation stems from two fundamental conflicts in the current training paradigm:
\begin{itemize}
    \item \textbf{Knowledge Sparsity of Linguistic Priors:} General-purpose LLMs, primarily driven by normative pre-training corpora, lack sensitivity to the specific ``non-normative'' error distributions of learners, such as homophone misusage and function word redundancies. Without strong underlying grammatical constraints, models struggle to strike a precise balance between fluency and grammatical correctness.
    \item \textbf{Over-correction Bias in Generation:} The Maximum Likelihood Estimation (MLE) objective used in SFT encourages models to shift input sentences toward high-probability regions of their internal distribution. Consequently, models tend to perform unnecessary paraphrasing instead of precise correction when encountering correct or slightly deviant sentences. This behavior results in a high false-positive rate, which contradicts the core objective of minimal editing.
\end{itemize}

\begin{figure*}[!t]
    \centering
    \includegraphics[width=0.85\linewidth]{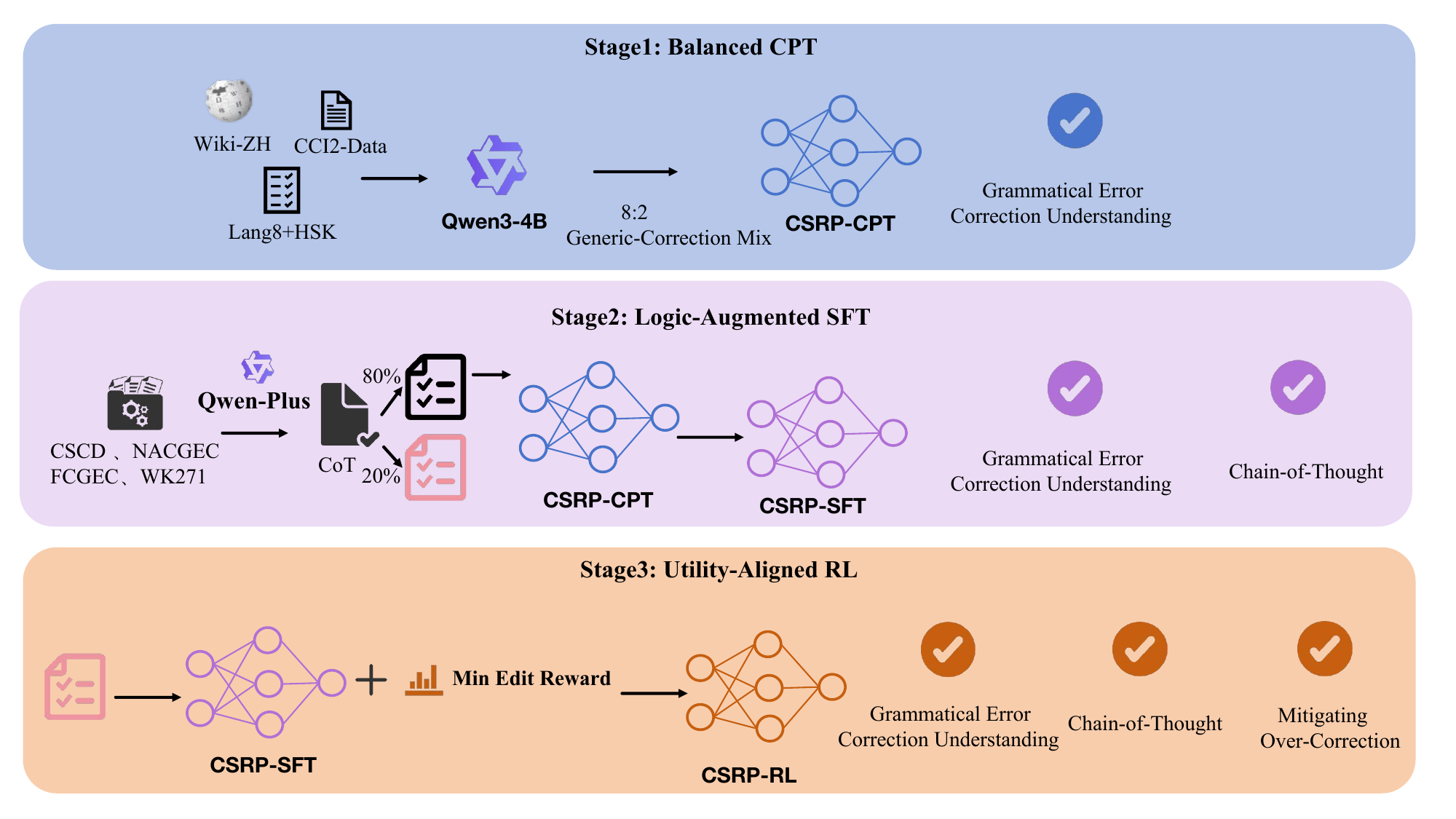}
    \caption{Overview of the proposed CSRP (CPT-SFT-RL) framework.}
    \label{fig:csrp_overview}
\end{figure*}
To address these challenges, as shown in Figure~\ref{fig:csrp_overview}, we propose the \textbf{CSRP (CPT-SFT-RL)} framework, a systematic pipeline designed to build a reliable correction system through knowledge internalization, rationale explicitization, and policy alignment:
\begin{itemize}
    \item \textbf{Phase I: Balanced Continued Pre-training (CPT).} We perform large-scale CPT on 5.9M samples using an 8:2 mixture ratio of general to correction-specific data. This stage internalizes fine-grained Chinese grammatical constraints into the parameter space, mitigating knowledge sparsity.
    \item \textbf{Phase II: Rationale-Augmented SFT.} We deviate from the traditional black-box mapping by distilling reasoning paths (Rationales) from high-performance teacher models. This Chain-of-Thought (CoT) mechanism guides the model to diagnose error types before executing corrections, enhancing transparency.
    \item \textbf{Phase III: Efficiency-Aware Policy Alignment.} To tackle the over-correction challenge, we introduce the \textbf{Group Relative Policy Optimization (GRPO)} algorithm. We design a multi-dimensional reward mechanism based on \textbf{Relative Improvement} and \textbf{Edit Efficiency}. Instead of blindly encouraging fluency, this mechanism explicitly rewards valid edits that reduce the distance to the target while penalizing stray modifications. Thus, the model learns to calibrate its decision boundaries, executing edits only when the corrective gain outweighs the fidelity cost.
\end{itemize}

\section{Related Work}

\subsection{Evolution of Correction Paradigms}
The field of Chinese text correction has undergone a significant transition from discriminative encoding to generative reconstruction. Early research in Chinese Spell Checking (CSC) focused on integrating phonological and visual constraints into BERT-based encoders, such as \textbf{SpellGCN} \cite{ji-etal-2021-spellbert}, \textbf{ReaLiSe} \cite{xu-etal-2021-realise}, and \textbf{PHMOSpell} \cite{huang-etal-2021-phmospell}. Subsequent works introduced specialized pre-training tasks and disentangled representations to mitigate phonetic-glyph confusion, including \textbf{PLOME} \cite{liu-etal-2021-plome}, \textbf{DORM-CSC} \cite{DORM-CSC2023}, and \textbf{PTCSpell} \cite{wei-etal-2023-ptcspell}.

In the era of Large Language Models (LLMs), the paradigm has shifted toward sequence-to-sequence rephrasing. \textbf{ReLM}\cite{liu-etal-2024-relm} re-conceptualized CSC as a language modeling task, while \textbf{C-LLM} \cite{li-etal-2024-c} explored character-by-character checking. For Chinese Grammatical Error Correction (CGEC), architectures have evolved from sequence-to-action models and syntax-enhanced frameworks like \textbf{SynGEC} \cite{zhang-etal-2022-syngec} to massive generative models such as \textbf{ChineseErrorCorrector3} \cite{ChineseErrorCorrector3}, which currently represents the state-of-the-art (SOTA). 

\subsection{Knowledge Acquisition and Domain Adaptation}
Effective CGEC requires capturing complex error distributions across diverse domains. Benchmarks such as  \textbf{NACGEC} \cite{NACGEC} established the foundation for evaluating native and learner-oriented texts. To bridge the gap between general pre-training and specialized correction, researchers have explored Retrieval-Augmented Generation (RAG). \textbf{MTCSC} \cite{MTCSC} and \textbf{RagID} \cite{COLING2025-RagID} utilize iterative refinement and few-shot retrieval to introduce external knowledge. Furthermore, \citet{ACL2025-307} demonstrated that multi-level structural cues (lexical and syntactic) are vital for precise error localization. Our work complements these by internalizing such priors via large-scale balanced Continued Pre-training (CPT), achieving spontaneous alignment of correction capabilities while avoiding the inference latency inherent in RAG systems \cite{Zhou2025-TrainingFree}.

\subsection{Reasoning, Reliability, and Policy Alignment}
The phenomenon of ``over-correction,'' where models produce hallucinatory edits, remains a critical bottleneck for practical application. Recent efforts have focused on enhancing model interpretability and controllability. \textbf{GEE} \cite{GEE} and \textbf{Rationale-based ICD} \cite{NAACL2025-ICD} utilize Chain-of-Thought (CoT) and explanatory information to guide the correction process. \textbf{ScholarGEC} \cite{ScholarGEC} further extends this to academic domains requiring extreme precision.

Despite these advancements, aligning LLMs with high-precision metrics like $F_{0.5}$ remains challenging due to the non-differentiable nature of GEC objectives. Traditional SFT based on Maximum Likelihood Estimation (MLE) often fails to calibrate the ``edit-or-not'' decision boundary. Our framework addresses this by leveraging \textbf{Group Relative Policy Optimization (GRPO)} to explicitly optimize for \textbf{relative improvement} and \textbf{edit efficiency}. This approach moves beyond simple score-chasing, aligning the model's policy with the linguistic principle of ``minimal intervention'' and effectively mitigating the over-correction bias observed in prior LLM-based systems \cite{MSLLM2025, GECFramework2024}.

\section{Methodology}

\begin{figure}[!t]
    \centering
    \includegraphics[width=1.0\linewidth]{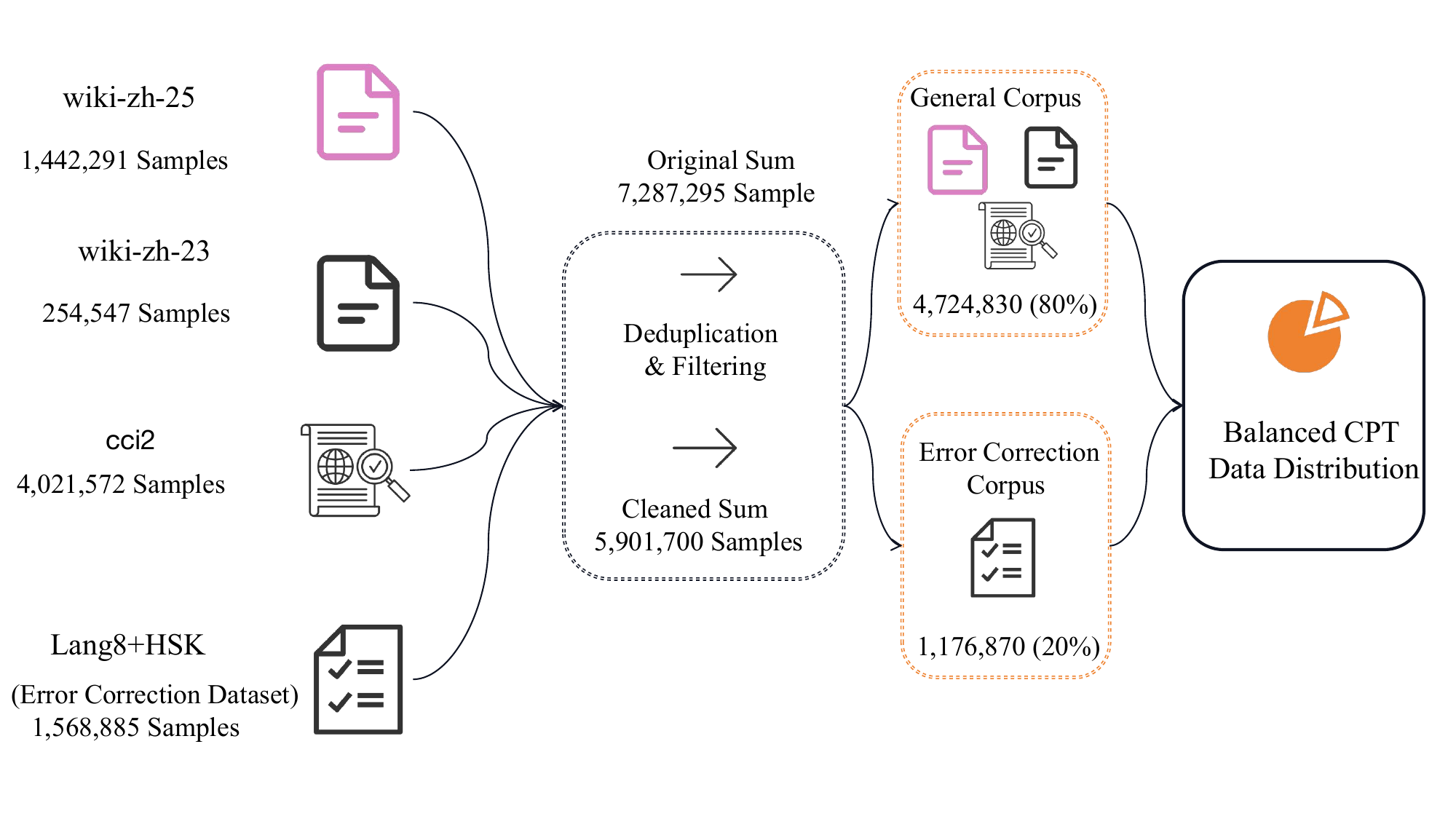}
    \caption{CPT Data Processing Process.}
    \label{fig:cpt_example}
\end{figure}

In this section, we present the \textbf{CSRP} framework, a systematic three-stage paradigm designed to transition a general-purpose Large Language Model (LLM) into a high-precision Chinese Grammatical Error Correction (CGEC) system. The pipeline evolves through: (i) \textbf{Balanced Continued Pre-training} for knowledge internalization; (ii) \textbf{Rationale-Augmented SFT} for diagnostic reasoning; and (iii) \textbf{Efficiency-Aware Policy Alignment} for decision boundary calibration.

\subsection{Phase I: Balanced Continued Pre-training}
Standard LLMs often exhibit \textit{knowledge sparsity} regarding the specific error distributions of learners. To internalize linguistic priors, we perform Continued Pre-training (CPT) on a refined 5.9M sample corpus.

\subsubsection{Data Refinement and Statistics}
We curate a comprehensive dataset $\mathcal{D}_{CPT}$ from four primary sources: \textit{wiki-zh-25}, \textit{wiki-zh-23}, \textit{cci2}, and \textit{lang8\cite{Lang8}+HSK\cite{HSK}}. As illustrated in Figure~\ref{fig:cpt_example}, we implement a rigorous refinement pipeline involving MinHash-based de-duplication and heuristic quality filtering. This process distilled the raw data from 7,287,295 to 5,901,700 high-quality samples. Notably, \textit{wiki-zh-23} was excluded due to extreme redundancy, while the \textit{cci2} and \textit{lang8+HSK} subsets were pruned to ensure high-fidelity linguistic patterns.

\subsubsection{General-to-Correction Balanced Mixture}
To mitigate \textit{catastrophic forgetting} of general reasoning capabilities, we adopt a \textbf{Balanced Mixture Strategy}. Inspired by the domain-specific pre-training insights in \citet{wen2023}, we employ an 8:2 ratio between general and correction-specific data. This translates to approximately 4.72M general samples (from \textit{wiki-zh-25} and \textit{cci2}) and 1.18M correction samples (from \textit{lang8+HSK}). The objective is to minimize the negative log-likelihood:
\begin{equation}
\mathcal{L}_{CPT}(\theta) = -\mathbb{E}_{x \sim \mathcal{D}_{CPT}} \left[ \sum_{t} \log P_{\theta}(x_t | x_{<t}) \right]
\end{equation}

\begin{figure}[!t]
    \centering
    \includegraphics[width=1.0\linewidth]{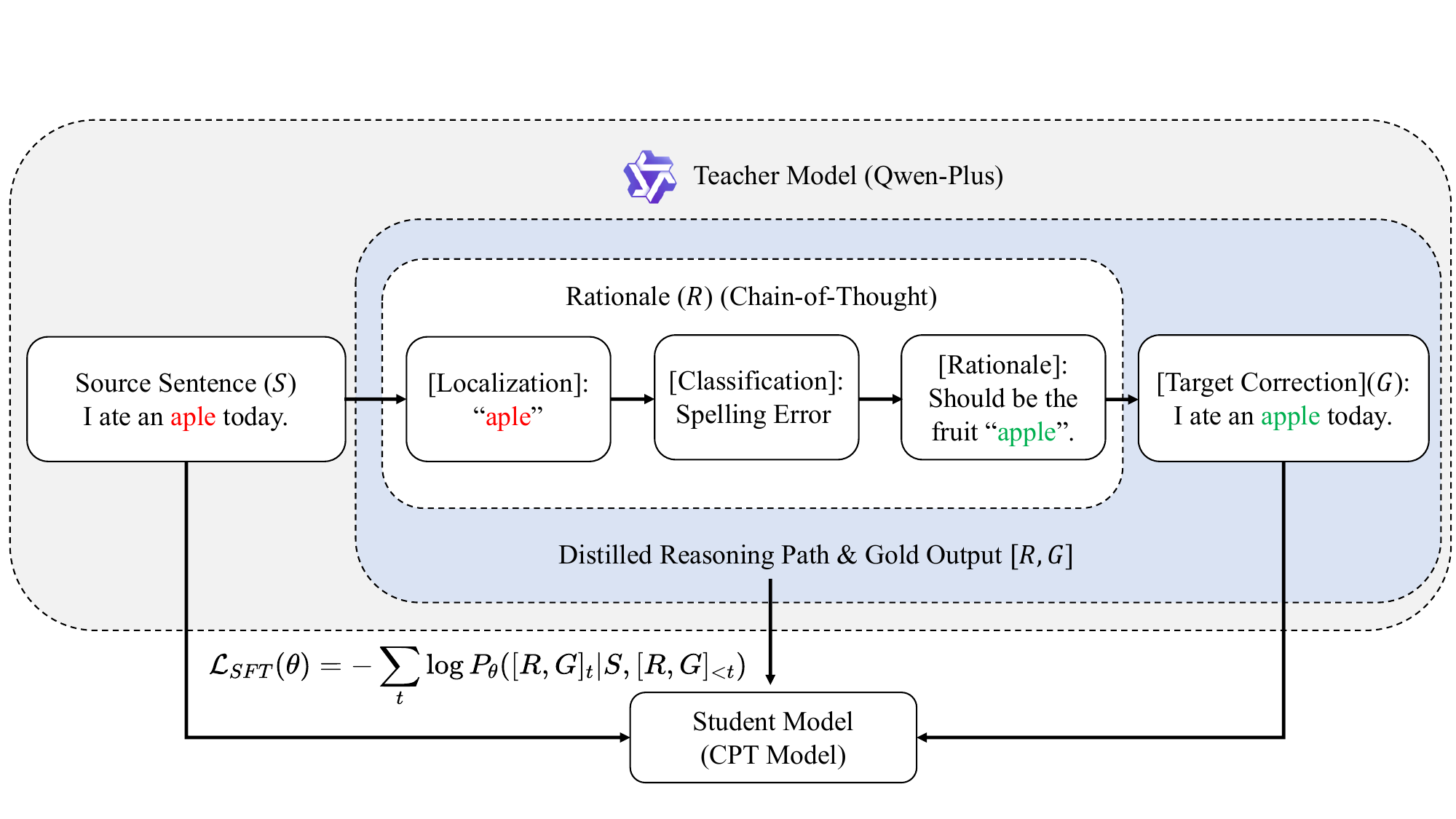}
    \caption{Cot Processing Process.}
    \label{fig:sft_cot}
\end{figure}

\subsection{Phase II: Rationale-Augmented SFT}
To transition from knowledge acquisition to active diagnosis, we introduce \textbf{Rationale-Augmented SFT}. Instead of a black-box mapping $S \rightarrow G$, we distill reasoning paths $R$ from a high-performance teacher model (Qwen-Plus) to guide the student's cognitive process.
 
Each rationale follows a structured Chain-of-Thought (CoT) format: \textit{[Localization] $\rightarrow$ [Classification] $\rightarrow$ [Rationale]}, as shown in Figure~\ref{fig:sft_cot}. The student model optimizes:
\begin{equation}
\mathcal{L}_{SFT}(\theta) = -\sum_{t} \log P_{\theta}([R, G]_t | S, [R, G]_{<t})
\end{equation}
This ``diagnose-before-correct'' paradigm ensures that the model's edits are grounded in explicit linguistic logic.
 
\noindent\textbf{Faithfulness of Distilled Rationales}

A critical concern in knowledge distillation is the reliability of teacher-generated reasoning paths, particularly the risk of hallucinated or templated explanations. We address this through a two-tiered quality control mechanism:

\begin{itemize}
    \item \textbf{Data level}: We apply strict filtering by discarding rationales that fail to follow the prescribed \texttt{<think>...</think>} format. Notably, Qwen-Plus itself exhibits a severe over-correction tendency when used directly as a corrector; this is precisely why its role is strictly limited to generating intermediate \emph{reasoning paths} between the fixed source $S$ and gold-standard target $G$, rather than final corrections.
    
    \item \textbf{Evaluation level}: A double-blind human study conducted by three annotators with relevant domain expertise on 1,000 randomly sampled rationales confirmed that 95.2\% are linguistically faithful (Cohen's $\kappa = 0.81$). This demonstrates high inter-annotator agreement and near-perfect rationale reliability.
\end{itemize}

Full annotation protocols and quality control details are provided in the Appendix.~\ref{sec:cot_quality}.

\subsection{Phase III: Efficiency-Aware Policy Alignment}
The most persistent challenge in CGEC is \textbf{over-correction}, where the model unnecessarily paraphrases correct prose. We address this by calibrating the decision boundary through \textbf{GRPO}.

\subsubsection{Mathematical Reward Modeling}
We define an \textbf{Efficiency-Aware Reward (EAR)} function. Let $S, P, G$ be the source, prediction, and ground-truth, and $d(\cdot, \cdot)$ be the Levenshtein distance. We derive two core metrics:

\noindent \textbf{(1) Relative Improvement ($RI$):}
\begin{equation}
RI = \frac{d(S, G) - d(P, G)}{d(S, G) + \epsilon}
\end{equation}
\textbf{(2) Edit Efficiency Ratio ($\eta$):} This serves as our fidelity constraint, penalizing excessive edits relative to the improvement gained:
\begin{equation}
\eta = \frac{d(S, G) - d(P, G)}{d(S, P) + \epsilon}
\end{equation}
where $\epsilon = 10^{-6}$. High $\eta$ signifies ``surgical precision.''

\subsubsection{Reward Function Formulation}
The reward $R_{EAR}$ provides distinct signals to align the model with the principle of minimal intervention:
\begin{equation}
R_{EAR} = 
\begin{cases} 
10.0, & \text{if } RI > 0.99 \\
2.0 + 5 RI \cdot \max(0, \eta), & \text{if } RI > 0 \\
-3.0, & \text{if } RI \leq 0 \\
-2.0, & \text{if } P = \emptyset
\end{cases}
\end{equation}

For source sentences that are already correct ($d(S, G) = 0$), the above cases reduce to a binary signal: identity mapping ($d(S, P) = 0$, i.e., $RI=0$) receives $+2.0$, rewarding the model for preserving well-formed text; any edit ($d(S, P) > 0$, which yields $RI \leq 0$) receives $-2.0$, directly penalizing over-correction on correct inputs. This explicit signal is the primary driver of the false positive rate reduction observed after RL training.

\subsubsection{Policy Optimization via GRPO}
GRPO optimizes the policy by contrasting $N$ completions $\{P_1, \dots, P_N\}$ for each prompt $S$:
\begin{equation}
\begin{aligned}
\mathcal{J}(\theta) = \frac{1}{N} \sum_{i=1}^{N} \bigg[ & \frac{R_i - \bar{R}}{\sigma_R} \log \pi_{\theta}(P_i | S) \\
& - \beta \mathbb{D}_{KL}(\pi_{\theta} || \pi_{ref}) \bigg]
\end{aligned}
\end{equation}
where $\bar{R}$ and $\sigma_R$ are the group reward mean and standard deviation. This mechanism encourages the model to prefer corrections with the highest \textbf{edit-to-improvement} ratio, effectively curbing the over-correction bias.

\section{Experiments}

In this section, we conduct extensive experiments to evaluate the effectiveness of the \textbf{CSRP} framework. We focus on two primary tasks: Grammatical Error Correction (GEC) on native-speaker texts and Chinese Spelling Check (CSC). Through comprehensive comparisons with state-of-the-art baselines and systematic ablation studies, we demonstrate that our three-stage curriculum learning approach achieves superior performance while maintaining edit efficiency.

\subsection{Experimental Setup}
\subsubsection{Data Setup}
To bridge the knowledge gap and align the model with correction objectives, we construct a three-stage data curriculum (CPT$\rightarrow$SFT$\rightarrow$RL).
Table~\ref{tab:dataset_stats} provides an overview of the data used at each stage.

\begin{table}[!htbp]
\centering
\small
\begin{tabular}{llr}
\hline
\textbf{Phase} & \textbf{Source} & \textbf{Count} \\ \hline
\multirow{2}{*}{CPT} & Wiki-zh, CCI2 & 4.7M \\
                    & Lang8, HSK    & 1.2M \\ \hline
\multirow{3}{*}{SFT} & CSCD-train, NACGEC & 37.5K \\
                     & FCGEC, CGED        & 55K \\
                     & WK271              & 251K \\
                     & \textit{(80\% split)} & \textit{269K} \\ \hline
RL   & SFT hold-out split (20\%) & 67K \\ \hline
\multirow{2}{*}{Eval} & NACGEC    & 5.8K \\
                      & CSCD-test & 5.0K \\ \hline
\end{tabular}
\caption{Data statistics for the three-stage training curriculum and 
evaluation sets. The SFT and RL phases use an 80:20 split of 
the correction data (total 336K samples after filtering). 
Note that the actual sample count (336K) is lower than the 
sum of individual subsets, as a portion of sentences 
containing sensitive content were rejected by the cloud-based 
teacher model's safety policy during CoT distillation.}
\label{tab:dataset_stats}
\end{table}

\noindent\textbf{Phase I (CPT).}
We curate a balanced corpus of 5.9M samples by mixing general-domain text (e.g., Wiki-zh, CCI2) and correction-specific data (e.g., Lang8, HSK\cite{HSK}) at an 8:2 ratio, aiming to internalize Chinese linguistic priors.

\noindent\textbf{Phase II (SFT).}
We aggregate supervision from multiple benchmarks (e.g., CSCD\cite{CSCD}, NACGEC\cite{NACGEC}, FCGEC\cite{xu-etal-2022-fcgec}, CGED\cite{CGED}, and WK271\cite{wk271}), yielding a total of 336K correction samples.
Each sample is augmented with a structured Chain-of-Thought (CoT) rationale distilled from a cloud-based teacher model (Qwen-Plus), which encourages the model to \emph{diagnose-before-correct} and improves interpretability.
As the teacher model's content safety policy rejects sentences containing sensitive content during distillation, along with standard data cleaning (e.g., deduplication), the final corpus is reduced to 336K successfully distilled samples.
Following this filtering, we apply an 80:20 split, allocating 269K samples (80\%) for supervised fine-tuning and reserving 67K samples (20\%) for the subsequent RL phase.

\noindent\textbf{Phase III (RL).}
We use the reserved 67K samples (20\% hold-out split) as prompts for policy optimization via GRPO, while the model retains knowledge from the 269K samples learned during supervised training.
This separation ensures that RL optimization explores beyond the supervised demonstrations while maintaining a strong foundation from SFT.

Detailed data sources, preprocessing pipelines, and dataset distributions are provided in Appendix~\ref{sec:app_data}.

\subsubsection{Baseline Models}
We compare CSRP with representative baselines spanning (i) PLM-based discriminative correction models, (ii) sequence-to-sequence Chinese GEC models, and (iii) general LLM prompting. We include strong recent systems such as HW-CGEC, ScholarGEC-14B, and ChineseErrorCorrect3-4B; details are in \textbf{Appendix~\ref{sec:baseline model}}.

\subsubsection{Evaluation Metrics}
\noindent \textbf{Evaluation.}
For GEC, we use the ChERRANT scorer~\cite{zhang-etal-2023-nasgec} and report Precision (P), Recall (R), and $F_{0.5}$, with $F_{0.5}$ as the primary metric since it emphasizes precision and penalizes over-corrections.
For CSC, we report character-level correction F1 following standard protocols~\cite{CSCD}.
Complete metric definitions, scoring procedures, and implementation details are provided in Appendix~\ref{sec:evaluation metrics}.

\subsubsection{Implementation Details}
\noindent \textbf{Training setup.}
We train CSRP with a three-stage pipeline: continued pre-training (CPT), supervised fine-tuning (SFT), and GRPO.
In GRPO, we regularize against the SFT initialization via a KL penalty and optimize a weighted reward that jointly considers correction quality and efficiency.
All hyperparameters (e.g., learning rates, batch sizes, reward weights), optimization settings, and compute details are reported in Appendix~\ref{sec:implementation}.

\subsection{Main Results}

\subsubsection{Performance on NACGEC}
Table~\ref{tab:nacgec_results} presents the main results on the NACGEC grammatical error correction benchmark. Our CSRP-4B model achieves a new state-of-the-art $F_{0.5}$ score of \textbf{50.99}, representing substantial improvements over all baseline systems.

\begin{table}[t]
\centering
\small
\begin{tabular}{l|ccc}
\hline
\textbf{Model} & \textbf{P} & \textbf{R} & \textbf{$F_{0.5}$} \\ \hline

BART\cite{BART} & 34.67 & 41.88 & 35.91 \\
HW-CGEC\cite{HW-CGEC} & 50.95 & 32.29 & 45.26 \\
ScholarGEC (14B)\cite{ScholarGEC} & 45.08 & \textbf{59.33} & 47.35 \\
CEC3 (4B)\cite{ChineseErrorCorrector3} & 54.20 & 34.75 & 48.74 \\ \hline

\textbf{CSRP (4B)} & \textbf{57.17} & 35.60 & \textbf{50.99} \\ \hline
\end{tabular}
\caption{Main results on the NACGEC benchmark.}
\label{tab:nacgec_results}
\end{table}

Compared to previous best results, CSRP achieves:
\begin{itemize}[leftmargin=*]
    \item \textbf{+2.25 points} over CEC3 (4B), the previous state-of-the-art 4B-scale model (48.74 $\rightarrow$ 50.99)
    \item \textbf{+3.64 points} over ScholarGEC (14B), despite using less than one-third of the parameters (47.35 $\rightarrow$ 50.99)
    \item \textbf{+5.73 points} over HW-CGEC, another strong specialized GEC system (45.26 $\rightarrow$ 50.99)
\end{itemize}

Notably, our model achieves the highest precision (\textbf{57.17}) among all compared systems, demonstrating superior correction accuracy. This high precision indicates that CSRP effectively avoids false positives -- incorrectly modifying text that is already correct. The precision advantage over CEC3 (+2.97 points) and ScholarGEC (+12.09 points) is particularly significant, confirming that our Efficiency-Aware Reward successfully suppresses the ``over-correction bias'' that commonly afflicts maximum likelihood estimation (MLE) trained models.

While our recall (35.60) is comparable to other high-precision systems like HW-CGEC (32.29) and CEC3 (34.75), it is notably lower than ScholarGEC's 59.33. This trade-off is intentional and aligned with the minimal-editing requirement in native-speaker correction: the $F_{0.5}$ metric explicitly prioritizes precision over recall (with a 2.5:1 weighting), reflecting the practical reality that conservative, high-confidence corrections are preferable to aggressive corrections that risk introducing new errors.

The BART baseline, despite being a strong sequence-to-sequence foundation, achieves only 35.91 $F_{0.5}$, highlighting the importance of specialized training curricula and reward-guided optimization for this task. The 15.08-point gap between BART and our model demonstrates that effective error correction requires more than general sequence transduction capabilities -- it demands carefully designed training strategies that balance linguistic knowledge, error pattern recognition, and editing conservatism.

\subsubsection{Performance on CSCD}
Table~\ref{tab:cscd_results} shows the  spelling check performance on the CSCD dataset. CSRP achieves a correction F1 score of \textbf{59.61}.

\begin{table}[t]
\centering
\small
\begin{tabular}{l|c}
\hline
\textbf{Model} & \textbf{F1} \\ \hline
BERT\cite{BERT_CSC} & 25.49 \\
SoftMask\cite{SoftMask} & 44.48 \\
SMBERT\cite{SMBERT} & 44.67 \\
MDCSpell+ARM\cite{liu-etal-2024-relm} & 48.93 \\
PGT (BERT)\cite{PGT} & 48.57 \\
GPT-4\cite{GPT4} & 54.41 \\ \hline
\textbf{CSRP (4B)} & \textbf{59.61} \\ \hline
\end{tabular}
\caption{Chinese spelling check performance on the CSCD dataset.}
\label{tab:cscd_results}
\end{table}

Key performance highlights include:
\begin{itemize}[leftmargin=*]
    \item \textbf{+5.20 points} over GPT-4 (54.41 $\rightarrow$ 59.61), demonstrating that our specialized training approach outperforms even powerful general-purpose models
    \item \textbf{+10.68 points} over MDCSpell+ARM (48.93 $\rightarrow$ 59.61), the best-performing discriminative baseline
    \item \textbf{+34.12 points} over BERT (25.49 $\rightarrow$ 59.61), highlighting the substantial gap between basic PLMs and our curriculum-trained model
\end{itemize}

The results indicate that CSRP's combination of balanced knowledge injection (Phase I) and edit-efficiency alignment (Phase III) significantly enhances the model's sensitivity to fine-grained phonetic and visual character substitutions. Character-level spelling errors are particularly challenging because they require: (i) deep understanding of phonetic similarities, (ii) recognition of visual similarities (e.g., characters with similar stroke patterns), and (iii) contextual semantic understanding to determine whether a character is appropriate in context.

The strong performance gap between GPT-4 and CSRP (+5.20 points) is particularly noteworthy. Despite GPT-4's massive scale and extensive pre-training, it underperforms our 4B-parameter specialized model, validating our hypothesis that task-specific curriculum learning and reinforcement-based alignment are more effective than scale alone for this specialized correction task. This suggests that the combination of domain-focused continual pre-training and efficiency-aware policy optimization provides advantages that cannot be easily achieved through general pre-training or in-context learning.

The discriminative models (BERT, SoftMask, SMBERT) achieve substantially lower scores (25.49-44.67 F1), likely due to their limited capacity to model long-range dependencies and generate corrections in an autoregressive manner. These models rely primarily on local context and character-level features, whereas our generative approach can leverage broader discourse context and learned linguistic patterns to make more informed correction decisions.

\subsection{Ablation Studies}
\subsubsection{Impact of Each Training Stage}
 
To systematically understand the contribution of each component, we conduct a progressive ablation study covering all three training stages. Table~\ref{tab:ablation_stages} reports Precision (P), Recall (R), NACGEC $F_{0.5}$, and CSCD F1 for each configuration.
 
\begin{table*}[t]
\centering
\small
\setlength{\tabcolsep}{6pt}
\begin{tabular}{l|cc|c|c}
\hline
\multirow{2}{*}{\textbf{Configuration}} &
  \multicolumn{2}{c|}{\textbf{NACGEC}} &
  \textbf{NACGEC} & \textbf{CSCD} \\
 & \textbf{P} & \textbf{R} & \textbf{$F_{0.5}$} & \textbf{F1} \\
\hline
SFT only (merged data)      & 42.13 & 34.02 & 40.21 & 49.71 \\
SFT + GRPO (w/o CPT)        & 50.54 & 33.75 & 45.97 & 52.96 \\
\hdashline
CPT + SFT (no CoT)          & 44.90 & 35.50 & 42.64 & 52.01 \\
CPT + SFT                   & 48.73 & 35.80 & 45.45 & 56.28 \\
CPT + SFT (w/ RL data)      & 52.20 & \textbf{36.00} & 47.21 & 57.92 \\
\textbf{Full CSRP}          & \textbf{57.17} & 35.60 & \textbf{50.99} & \textbf{59.61} \\
\hline
\end{tabular}
\caption{Progressive ablation study. ``SFT only (merged data)'' combines all data without CPT. ``SFT + GRPO (w/o CPT)'' applies GRPO directly to the base model to isolate the RL contribution. ``CPT + SFT (no CoT)'' excludes chain-of-thought reasoning. ``CPT + SFT (w/ RL data)'' uses all data for SFT as a controlled baseline.}
\label{tab:ablation_stages}
\end{table*}
 
The results reveal a clear progression of capabilities and allow us to disentangle the contribution of each component.
 
\noindent \textbf{(1) Importance of CPT Stage.}
Comparing ``SFT only (merged data)'' (40.21 $F_{0.5}$ / 49.71 F1) with ``CPT + SFT'' (45.45 / 56.28) demonstrates the critical role of continual pre-training. The CPT stage provides domain-specific linguistic knowledge that cannot be obtained by simply merging all correction data into supervised fine-tuning, yielding improvements of +5.24 $F_{0.5}$ and +6.57 F1.
Crucially, CPT and RL contribute through different, complementary mechanisms: CPT raises the performance \emph{baseline} by internalizing fine-grained Chinese linguistic priors (phonetic patterns, function-word constraints, etc.), whereas RL refines the \emph{editing policy} by calibrating decision boundaries. Neither stage can substitute for the other. Specifically, ``SFT + GRPO (w/o CPT)'' achieves 45.97 $F_{0.5}$, nearly matching the upper bound of pure CPT+SFT (45.45) but still lagging behind the full pipeline by 5.02 points, confirming that domain-adaptive pre-training provides linguistic grounding that RL exploration alone cannot recover.
 
\noindent \textbf{Disentangling CPT and RL contributions.}
The ``SFT + GRPO (w/o CPT)'' ablation is critical for attributing gains correctly. As shown in Table~\ref{tab:ablation_stages}, applying GRPO directly to the SFT model (bypassing CPT entirely) yields a precision boost of +8.41 points and $F_{0.5}$ gain of +5.76 over the SFT-only baseline. This mirrors almost exactly the RL contribution in the full pipeline (+8.44 P / +5.54 $F_{0.5}$). Two conclusions follow. First, the Efficiency-Aware Reward independently and effectively calibrates the ``edit-or-not'' decision boundary, regardless of pre-training scale; its benefit is not merely a downstream consequence of the larger CPT corpus. Second, CPT and RL operate via \emph{orthogonal mechanisms}: CPT encodes \emph{what} constitutes a grammatical error in Chinese by internalizing the linguistic distribution of the correction domain, while RL encodes \emph{when} to intervene by optimizing the efficiency of edits against a reward signal. Because the two stages correct for different failure modes (knowledge sparsity vs.\ over-correction bias), their gains are additive rather than redundant, and both are jointly necessary for the full CSRP pipeline.
 
\noindent \textbf{(2) Importance of Distilled CoT in SFT Stage.}
The comparison between ``CPT + SFT (no CoT)'' (42.64 / 52.01) and ``CPT + SFT'' (45.45 / 56.28) quantifies the value of teacher-distilled Chain-of-Thought rationales. Incorporating structured reasoning paths (\textit{[Localization] $\rightarrow$ [Classification] $\rightarrow$ [Rationale]}) improves performance by +2.81 $F_{0.5}$ and +4.27 F1. The gains are particularly pronounced on CSCD (+4.27 F1), where fine-grained phonetic/visual diagnosis is critical. By requiring the model to \emph{diagnose-before-correct}, the CoT objective injects structured error-type knowledge that is difficult to acquire from correction pairs alone. Furthermore, on test samples requiring complex multi-span edits (gold edit distance $> 3$), CSRP maintains a Recall of 31.2\%, significantly outperforming the SFT-only baseline (24.5\%), confirming that faithful CoT supervision equips the model with genuine diagnostic capability beyond shallow pattern memorization.
 
\noindent \textbf{(3) Contribution of RL Stage.}
To isolate the RL contribution beyond data quantity, we compare ``CPT + SFT (w/ RL data)'' (47.21 / 57.92) against ``Full CSRP'' (50.99 / 59.61), both using identical amounts of training data. The RL stage provides a further gain of +3.78 $F_{0.5}$ and +1.69 F1, validating that the Efficiency-Aware Reward introduces a qualitatively different optimization signal rather than simply benefiting from more supervised samples.
As shown by the precision columns, the RL stage is the primary driver of precision improvement: it pushes precision from 52.20 (CPT+SFT w/ RL data) to 57.17 (Full CSRP), a gain of +4.97 points, while recall remains stable (36.00 $\rightarrow$ 35.60). This asymmetric impact confirms that RL specifically calibrates the ``edit-or-not'' decision boundary rather than uniformly reducing interventions.

\subsubsection{Impact of Reinforcement Learning on Precision-Recall Trade-off}
To understand how RL alignment affects model behavior, we analyze the precision-recall trade-off before and after the GRPO stage. Table~\ref{tab:ablation_precision} presents detailed results.

\begin{table}[h]
\centering
\small
\begin{tabular}{l|ccc}
\hline
\multicolumn{4}{c}{\textbf{NACGEC}} \\ \hline
\textbf{Configuration} & \textbf{P} & \textbf{R} & \textbf{$F_{0.5}$} \\ \hline
CPT + SFT & 48.73 & 35.80 & 45.45 \\
CPT + SFT + GRPO & 57.17 & 35.60 & 50.99 \\ \hline
\textbf{Change} & \textcolor{blue}{+8.44} & \textcolor{red}{-0.20} & \textcolor{blue}{+5.54} \\ \hline
\end{tabular}

\vspace{0.3cm}

\begin{tabular}{l|ccc}
\hline
\multicolumn{4}{c}{\textbf{CSCD}} \\ \hline
\textbf{Configuration} & \textbf{P} & \textbf{R} & \textbf{F1} \\ \hline
CPT + SFT  & 58.85 & 53.92 & 56.28 \\
CPT + SFT + GRPO & 66.22 & 53.20 & 59.61 \\ \hline
\textbf{Change} & \textcolor{blue}{+7.37} & \textcolor{red}{-0.72} & \textcolor{blue}{+3.33} \\ \hline
\end{tabular}
\caption{Impact of GRPO on precision-recall trade-off. The RL stage significantly improves precision while maintaining comparable recall, demonstrating effective learning of conservative editing behavior.}
\label{tab:ablation_precision}
\end{table}

The results reveal a crucial insight into how GRPO reshapes model behavior:

\noindent \textbf{Substantial Precision Gains:} After RL training, precision increases dramatically on both tasks (+8.44 points on NACGEC, +7.37 points on CSCD). This represents a relative improvement of 17.3\% and 12.5\% respectively. The large precision gains indicate that the Efficiency-Aware Reward successfully teaches the model to avoid unnecessary edits and reduce false positives (over-correction).

\noindent \textbf{Maintained Recall:} Crucially, recall decreases only marginally (-0.20 on NACGEC, -0.72 on CSCD), representing less than 1-2\% relative change. This demonstrates that GRPO does not simply make the model more conservative by reducing all corrections; rather, it helps the model distinguish between necessary and unnecessary edits. The model learns to be selective, proposing corrections only when confident they are warranted.

\noindent \textbf{Optimal Precision-Recall Balance:} The asymmetric impact on precision vs. recall aligns perfectly with the goals of native-speaker correction. The SFT model, trained with maximum likelihood estimation, tends toward over-correction to maximize the likelihood of matching reference corrections. In contrast, the GRPO model, guided by the Efficiency-Aware Reward, learns that proposing fewer but more accurate corrections yields higher overall reward. This shift from ``correct liberally'' to ``correct conservatively'' is precisely what the minimal-editing principle demands.

\noindent \textbf{Consistency Across Tasks:} The pattern holds consistently across both GEC (NACGEC) and CSC (CSCD) tasks. On NACGEC, the precision boost (+8.44) far exceeds the recall drop (-0.20), yielding a strong net positive on the precision-weighted $F_{0.5}$ metric (+5.54). On CSCD, similar dynamics apply: precision gains (+7.37) vastly outweigh recall losses (-0.72), resulting in substantial F1 improvements (+3.33). This consistency suggests that the efficiency-aware training signal generalizes across different error types and correction paradigms.

\subsubsection{Why Does RL Improve Precision Without Sacrificing Recall?}
Three complementary mechanisms explain this outcome.
\textbf{Confidence calibration}: GRPO's group-relative comparison ($N=8$ candidates) develops better-calibrated estimates, so the model proposes edits only when candidates consistently agree on high rewards.
\textbf{Dual editing signal}: Unlike MLE, the Efficiency-Aware Reward explicitly penalizes changes to correct text, teaching the model both ``what to correct'' and ``what to preserve.''
\textbf{Conservative strategy discovery}: Through reward-based learning, the model internalizes that conservative edits yield higher rewards on low-error inputs, suppressing unnecessary modifications without reducing valid corrections.
Together, these mechanisms yield substantial precision gains (+8.44/+7.37 points) at minimal recall cost (-0.20/-0.72 points).

\section{Conclusion}
We present CSRP, a three-stage framework progressing from linguistic knowledge internalization (CPT) to reasoning-augmented correction (CoT-SFT) to efficiency-aware policy alignment (GRPO with EAR). Our 4B-parameter model sets new state-of-the-art results on NACGEC (50.99 $F_{0.5}$) and CSCD (59.61 F1), surpassing both larger models (14B) and GPT-4, while substantially reducing over-correction (+8.44/+7.37 precision gains with negligible recall loss). A controlled ablation applying GRPO without CPT (45.97 $F_{0.5}$) confirms that the two stages contribute orthogonally and are jointly necessary. Our findings demonstrate that principled curriculum design and efficiency-aware optimization outperform both scale and data quantity alone. Future work includes document-level correction, interactive refinement, and cross-lingual transfer.

\section{Limitations}
While our CSRP framework demonstrates strong performance on Chinese spelling correction, several limitations warrant discussion:

\noindent \textbf{(1) Dependency on Teacher Model Quality.} 
The chain-of-thought reasoning in our SFT stage relies on distillation from a teacher model (Qwen-Plus). The quality and diversity of generated rationales are bounded by the teacher's capabilities. Errors or biases in teacher-generated explanations may propagate to the student model, potentially affecting correction interpretability. We mitigate this risk through strict edit-distance filtering of teacher outputs and a human validation study (Appendix~\ref{sec:cot_quality}), but residual bias cannot be fully excluded.

\noindent \textbf{(2) Computational Cost of RL Training.} 
GRPO requires generating multiple candidate outputs ($N=8$ per input) during training, increasing computational cost compared to standard supervised learning. While this cost is justified by performance gains, it may be prohibitive for researchers with limited resources. Reducing to $N=4$ yields only a marginal drop (50.99 $\rightarrow$ 50.61 $F_{0.5}$) while cutting RL sampling cost by 50\%, offering a practical efficiency trade-off.

\bibliography{custom}

\clearpage
\appendix

\section{Data Curriculum and Composition}
\label{sec:app_data}
We implement a three-stage data curriculum to progressively transition the model from general linguistic competence to specialized error-correction expertise.

\noindent \textbf{(1) Stage I: Continual Pre-Training (CPT, 5.9M Samples)} \\
To instill robust grammatical priors, we curate a 5.9M sample corpus for continual pre-training. Following the balanced mixture strategy (8:2 ratio) inspired by \citet{wen2023}, we combine 4.7M general-domain samples (\textit{wiki-zh-25},\textit{wiki-zh-23}, \textit{cci2}) with 1.2M correction-specific samples (\textit{Lang8+HSK\cite{HSK}}). The general-domain data provides broad linguistic knowledge covering diverse topics and writing styles, while the correction-specific data introduces the model to common error patterns and their corrections.

As shown in Table~\ref{tab:data_processing}, we performed rigorous data refinement to ensure training quality. Specifically, we implemented a multi-level data cleaning strategy: (1) \textbf{Length Filtering}: We retained texts between 5-4096 characters, filtering out samples that were too short or too long; (2) \textbf{Traditional Chinese Detection}: We used the OpenCC tool to detect and filter texts containing traditional Chinese characters, ensuring simplified Chinese consistency across the corpus; (3) \textbf{Hash-based Deduplication}: We removed exact duplicates through exact matching, ensuring the uniqueness of each text; (4) \textbf{Sentence Integrity Preservation}: For long texts, we performed splitting at natural delimiters such as periods and newlines (maximum length 512 characters), and applied punctuation normalization to segments longer than 128 characters. This refinement process distilled the raw 7.287M samples into a high-fidelity 5.902M pre-training set, with general-domain and correction-specific samples mixed at an 8:2 ratio (4.707M:1.177M), removing approximately 1.386M low-quality or duplicate samples (19.0\%).

\noindent \textbf{(2) Stage II: Supervised Fine-Tuning (SFT, 269K Samples)} \\
We aggregate a diverse instruction-tuning set comprising 269K samples across multiple error correction scenarios. As shown in Table~\ref{tab:sft_data_composition}, the data composition includes:
\begin{itemize}[leftmargin=*]
    \item \textbf{Chinese Spelling Check (CSC)} : CSCD-NS\cite{CSCD} (native-speaker errors), WK271\cite{wk271}+SIGHAN15\cite{sig15}, representing character-level spelling errors with phonetic and visual confusion patterns
    \item \textbf{Chinese Grammatical Error Correction (CGEC)}: CGED\cite{CGED} (learner errors), NACGEC\cite{NACGEC} (native-speaker errors), and FCGEC\cite{xu-etal-2022-fcgec} (fine-grained annotations), covering diverse grammatical error types including word selection, syntax, redundancy, and missing components
\end{itemize}

After removing duplicates, the raw 345K samples were refined to 339K, which were then processed through reasoning augmentation. 

\begin{table}[t]
\centering
\small
\begin{tabular}{l|rrr}
\hline
\textbf{Dataset} & \textbf{Train} & \textbf{Test} & \textbf{Type} \\ \hline
CSCD-NS & 30,000 & 5,000 & CSC \\
WK271+SIGHAN15 & 251,835 & 1,100 & CSC \\
CGED & 20,449 & - & CGEC \\
NACGEC & 7,568 & 5,869 & CGEC \\
FCGEC & 35,355 & - & CGEC \\ \hline
\textbf{Subtotal} & 345,207 & 11,969 & - \\
After Deduplication & 338,530 & 11,969 & - \\
After CoT Augmentation & 336,604 & 11,946 & - \\ \hline
\multicolumn{4}{l}{\footnotesize Final Split: SFT (80\%) = 269,283; RL (20\%) = 67,321} \\ \hline
\end{tabular}
\caption{Data composition for Stage II (SFT) and Stage III (RL). }
\label{tab:sft_data_composition}
\end{table}

\noindent \textbf{Chain-of-Thought Reasoning Augmentation:} \\
Crucially, each sample is augmented with \textbf{Reasoned Rationales} distilled from a teacher model (Qwen-Plus\cite{qwen3}) to provide explicit diagnostic guidance. These rationales follow a structured three-step format enclosed in \texttt{<think>...</think>} tags: (i) \textit{error type identification} -- categorizing the error, (ii) \textit{error analysis} -- pinpointing the problematic text spans and explaining the root cause, and (iii) \textit{correction justification} -- explaining why the proposed correction resolves the issue. 

Due to content safety policies of the commercial LLM API, approximately 0.57\% of samples (1,926 training samples and 23 test samples) could not be augmented and were excluded, resulting in 336,604 training samples and 11,946 test samples. This chain-of-thought supervision enables the model to learn not just \textit{what} to correct, but \textit{why} certain edits are warranted, thereby improving both correction accuracy and interpretability.

\noindent \textbf{(3) Stage III: Reinforcement Learning from Policy Optimization } \\
For the Group Relative Policy Optimization (GRPO) phase, we allocate 20\% of the augmented training data for policy learning. This split ensures sufficient diversity for exploring correction strategies while reserving the majority of data for supervised pre-training. For each sentence, we generate $N=8$ candidate corrections using nucleus sampling (temperature $T=1.0$). This multi-candidate setup allows the model to explore different correction strategies while being guided by the Efficiency-Aware Reward (EAR) signal, which explicitly penalizes unnecessary edits while rewarding valid corrections. The 80-20 split between SFT and RL phases ensures that the model first acquires solid correction capabilities through supervised learning before refining its policy through reinforcement learning.

\noindent \textbf{(4) Evaluation Datasets} \\
We evaluate our model on two benchmarks representing different aspects of Chinese text correction:
\begin{itemize}[leftmargin=*]
    \item \textbf{NACGEC} \cite{NACGEC}: A native-speaker grammatical error correction benchmark containing 5,869 test sentences spanning news articles, essays, and social media posts. Errors include word choice, grammar, redundancy, missing words, and word order issues.
    \item \textbf{CSCD} \cite{CSCD}: A Chinese spelling check dataset with 5000 test sentences containing naturally occurring character-level errors.
\end{itemize}

\begin{table}[t]
\centering
\small
\begin{tabular}{l|cc}
\hline
\textbf{Data Source} & \textbf{Raw Count} & \textbf{After Dedup \& Filter} \\ \hline
Wiki-zh-25 & 1,442,291 & 1,417,899 \\
Wiki-zh-23 & 254,547 & 0 \\
CCI2 & 4,021,572 & 3,306,931 \\
Lang8+HSK & 1,568,885 & 1,176,870 \\ \hline
\textbf{Total} & \textbf{7,287,295} & \textbf{5,901,700} \\
\multicolumn{3}{l}{\footnotesize General:Correction = 8:2 (4,707,480:1,176,870)} \\ \hline
\end{tabular}
\caption{Data processing statistics for the CPT phase.}
\label{tab:data_processing}
\end{table}

\section{Baseline Models}
\label{sec:baseline model}
We compare CSRP against representative models from three categories:

\noindent \textbf{(1) Discriminative/PLM-based Approaches:} 
\begin{itemize}[leftmargin=*]
    \item \textbf{BERT} \cite{BERT_CSC}: Pre-trained bidirectional encoder fine-tuned on correction tasks using masked language modeling
    \item \textbf{SoftMask} \cite{SoftMask}: Employs soft-masking mechanism to detect and correct errors
    \item \textbf{SMBERT} \cite{SMBERT}: Spelling correction model with semantic matching
    \item \textbf{MDCSpell+ARM} \cite{liu-etal-2024-relm}: Multi-task learning approach with auxiliary reading module
    \item \textbf{PGT (BERT)} \cite{PGT}: Prior knowledge-guided teacher network that uses distillation learning to reduce over-correction in PLM-based spelling correction models
\end{itemize}

\noindent \textbf{(2) Sequence-to-Sequence GEC Models:}
\begin{itemize}[leftmargin=*]
    \item \textbf{BART}\cite{BART}: Denoising autoencoder adapted for Chinese GEC
    \item \textbf{HW-CGEC} \cite{HW-CGEC}: Huawei's CGEC system that won first place in NLPCC2023 shared task. The system employs BART-based sequence-to-sequence architecture enhanced with data augmentation and curriculum learning strategies.
    \item \textbf{ScholarGEC-14B} \cite{ScholarGEC}: Large-scale (14B parameters) generative model specifically trained for Chinese academic writing correction
    \item \textbf{ChineseErrorCorrect3-4B (CEC3)} \cite{ChineseErrorCorrector3}: State-of-the-art 4B parameter model trained on extensive Chinese correction corpora
\end{itemize}

\noindent \textbf{(3) General Large Language Models:}
\begin{itemize}[leftmargin=*]
    \item \textbf{GPT-4}: We evaluate OpenAI's GPT-4\cite{GPT4} using few-shot prompting with 3 correction examples
\end{itemize}

\section{Evaluation Metrics}
\label{sec:evaluation metrics}

\noindent \textbf{For Grammatical Error Correction (GEC):} We adopt the widely-used \textbf{ChERRANT} (Chinese ERRANT) scorer \cite{zhang-etal-2023-nasgec}, a character-level evaluation toolkit adapted from the ERRANT framework for Chinese. ChERRANT performs character-level alignment between system outputs and gold-standard references, which alleviates evaluation inaccuracies caused by word segmentation errors in Chinese. We report three key metrics:
\begin{itemize}[leftmargin=*]
    \item \textbf{Precision (P)}: Proportion of system-proposed edits that match the reference corrections
    \item \textbf{Recall (R)}: Proportion of gold-standard errors that are successfully detected and corrected
    \item \textbf{$F_{0.5}$ Score}: Weighted harmonic mean that prioritizes precision over recall with a 2.5:1 ratio
\end{itemize}
We prioritize $F_{0.5}$ as the primary metric because it aligns with the minimal-editing principle in native-speaker correction: false positives (over-correction) are more detrimental than false negatives (under-correction), as they introduce unintended changes to originally correct text. The $F_{0.5}$ score is computed as:
\begin{equation}
F_{0.5} = (1 + 0.5^2)\,\frac{P \cdot R}{0.5^2 \cdot P + R}
\end{equation}

\noindent \textbf{For Chinese Spelling Check (CSC):} Following standard CSC evaluation protocols \cite{CSCD}, we report character-level \textbf{Correction F1 score}, which measures the model's ability to both detect error positions and provide correct replacements. This metric requires exact character-level matches for a correction to be counted as correct.

\section{Implementation Details}
\label{sec:implementation}

\noindent \textbf{Phase I (CPT):} We train for 3 epochs over the 5.9M sample corpus with a learning rate of $1 \times 10^{-4}$, using a cosine annealing schedule with 10\% warmup steps. We use a per-device batch size of 4 with 4 gradient accumulation steps. This phase is implemented using LLaMAFactory.

\noindent \textbf{Phase II (SFT):} We train for 3 epochs with a learning rate of $1 \times 10^{-4}$, using a cosine annealing schedule with 10\% warmup steps. We use a per-device batch size of 8 with 8 gradient accumulation steps. This phase is implemented using LLaMAFactory.

\noindent \textbf{Phase III (GRPO):} We employ group relative policy optimization with a group size of $N=8$ candidates per prompt. The KL divergence coefficient is set to $\beta=0.01$ to prevent the policy from deviating too far from the SFT initialization. The learning rate is $1 \times 10^{-5}$ with a cosine annealing schedule and 10\% warmup steps, and we train for 8 epochs with a per-device batch size of 32 and 2 gradient accumulation steps. This phase is implemented using the TRL (Transformer Reinforcement Learning) library.

All training is conducted on 4$\times$NVIDIA H800 (80GB) GPUs using DeepSpeed ZeRO-3 optimization with mixed-precision (bfloat16) training.



\section{Chain-of-Thought Data Construction}
\label{sec:appendix_cot}


To enhance the model's interpretability and reasoning capabilities, we augment our training data with structured Chain-of-Thought (CoT) rationales. This section describes the methodology for constructing reasoning-augmented supervision signals using Qwen-Plus as the teacher model.

\subsection{Rationale Generation Framework}

We employ Qwen-Plus to distill explicit reasoning paths for each correction example. The rationale generation follows a structured three-component format:

\begin{itemize}[leftmargin=*]
    \item \textbf{Error Type Identification:} Classify the error into predefined categories (spelling errors, word collocation errors, part-of-speech errors, word order errors, missing/redundant components, connective word misuse, ambiguous reference, semantic-logical inconsistency, or no error).
    \item \textbf{Correction Justification:} Explain why the identified text is erroneous and how the correction addresses the issue.
    \item \textbf{Corrected Output:} Provide the final corrected sentence.
\end{itemize}

\subsection{Prompting Strategy}

For each error-correction pair $(S, G)$ in our training corpus, we query Qwen-Plus with a carefully designed instruction. The prompt instructs the model to act as a professional correction expert, systematically analyzing input sentences and identifying error types from predefined categories.

The model responds in the following structured format:
\begin{verbatim}
<think>
Error Type: [error_category]
Correction Rationale: [detailed_explanation]
</think>
[corrected_sentence]
\end{verbatim}

\subsection{Illustrative Example}

To demonstrate the CoT construction process, we present a concrete example. The source sentence contains a spelling error where visually similar characters are confused.

\noindent \textbf{Input (Source Sentence):}
\begin{quote}
\begin{CJK}{UTF8}{gbsn}
我今天吃了一个平果。
\end{CJK}
\textit{(Literal: ``I today ate one ping-guo.''; Intended: ``I ate an apple today.'')}
\end{quote}

\noindent \textbf{Generated CoT Rationale (from Qwen-Plus):}

\begin{quote}
\small
\texttt{<think>}

\textbf{Error Type:} Spelling error (visually similar characters)

\textbf{Correction Rationale:} The original sentence uses ``\begin{CJK}{UTF8}{gbsn}平果\end{CJK}'' (ping-guo) which should be ``\begin{CJK}{UTF8}{gbsn}苹果\end{CJK}'' (apple). The characters ``\begin{CJK}{UTF8}{gbsn}平\end{CJK}'' (flat/level) and ``\begin{CJK}{UTF8}{gbsn}苹\end{CJK}'' (apple [character]) are visually similar in their left radical component, leading to the spelling error. ``\begin{CJK}{UTF8}{gbsn}平果\end{CJK}'' is not a valid lexical item in modern Chinese, whereas ``\begin{CJK}{UTF8}{gbsn}苹果\end{CJK}'' is the standard term for the fruit ``apple.'' The correction uses the proper character form to eliminate ambiguity.

\texttt{</think>}

\begin{CJK}{UTF8}{gbsn}
我今天吃了一个苹果。
\end{CJK}

\textit{(Translation: ``I ate an apple today.'')}
\end{quote}

This rationale explicitly identifies the error type (spelling error caused by visual similarity between character components), explains the linguistic consequence (invalid vs. valid lexical item), and provides the corrected output. Such structured reasoning guides the student model to learn not only \textit{what} to correct but also \textit{why} certain edits are linguistically necessary.

\subsection{Data Processing Pipeline}

The complete CoT data construction pipeline consists of four stages:

\begin{enumerate}
    \item \textbf{Sampling:} Extract the source sentence $S$ from our correction corpus.
    \item \textbf{Rationale Generation:} Query Qwen-Plus with the instruction template and source sentence, obtaining a structured response $R$ containing error analysis and correction.
    \item \textbf{Validation:} Verify that: (a) the generated rationale follows the prescribed format, (b) the corrected output matches or is semantically equivalent to the gold reference $G$, and (c) the explanation is linguistically sound and non-trivial.
    \item \textbf{Integration:} Integrate valid rationales into the training data, forming triplets $(S, R, G)$ where $R$ encodes the diagnostic reasoning path.
\end{enumerate}

\subsection{Quality Control}

To ensure high-quality rationales, we implement three filtering mechanisms:

\begin{itemize}[leftmargin=*]
    \item \textbf{Format Compliance:} Discard samples where the output does not adhere to the \texttt{<think>...<\textbackslash think>} structure or lacks clear error type and rationale fields.
    \item \textbf{Explanation Quality:} Filter out generic or tautological explanations (e.g., ``The original sentence has errors, the corrected sentence fixes them'') that provide no linguistic insight.
\end{itemize}

After applying these quality filters, we obtain approximately 314K high-quality reasoning-augmented samples for Supervised Fine-Tuning (SFT). This CoT supervision enables the model to develop transparent, principle-based correction strategies rather than opaque pattern matching, contributing to both performance gains and interpretability improvements demonstrated in our main results.

\section{Error Type Examples}
\label{sec:appendix_error_types}

This section presents representative examples for each of the nine error categories used in our Chain-of-Thought (CoT) rationale generation. All examples are drawn from real-world Chinese text correction scenarios and illustrate the linguistic phenomena that our model is trained to identify and correct.

\subsection{Spelling Error}

Spelling errors arise when characters are incorrectly substituted due to phonetic similarity (homophones) or visual similarity (glyph-level confusion).

\noindent \textbf{Example 1 -- Visual Similarity:}
\begin{itemize}[leftmargin=*]
    \item \textbf{Source:} \begin{CJK}{UTF8}{gbsn}我今天吃了一个平果。\end{CJK}
    \item \textbf{Error:} \begin{CJK}{UTF8}{gbsn}平果\end{CJK} $\rightarrow$ \begin{CJK}{UTF8}{gbsn}苹果\end{CJK} (visual similarity between \begin{CJK}{UTF8}{gbsn}平\end{CJK} and \begin{CJK}{UTF8}{gbsn}苹\end{CJK})
    \item \textbf{Corrected:} \begin{CJK}{UTF8}{gbsn}我今天吃了一个苹果。\end{CJK}
\end{itemize}

\noindent \textbf{Example 2 -- Phonetic Similarity:}
\begin{itemize}[leftmargin=*]
    \item \textbf{Source:} \begin{CJK}{UTF8}{gbsn}他在也不想见到她了。\end{CJK}
    \item \textbf{Error:} \begin{CJK}{UTF8}{gbsn}在\end{CJK} $\rightarrow$ \begin{CJK}{UTF8}{gbsn}再\end{CJK} (homophone confusion; both pronounced ``z\`ai'')
    \item \textbf{Corrected:} \begin{CJK}{UTF8}{gbsn}他再也不想见到她了。\end{CJK}
\end{itemize}

\subsection{Word Collocation Error}

Collocation errors occur when word combinations violate conventional or idiomatic usage patterns in Chinese.

\noindent \textbf{Example:}
\begin{itemize}[leftmargin=*]
    \item \textbf{Source:} \begin{CJK}{UTF8}{gbsn}他对这个问题进行了深厚的研究。\end{CJK}
    \item \textbf{Error:} \begin{CJK}{UTF8}{gbsn}深厚的研究\end{CJK} is a non-idiomatic collocation
    \item \textbf{Corrected:} \begin{CJK}{UTF8}{gbsn}他对这个问题进行了深入的研究。\end{CJK}
\end{itemize}

\subsection{Part-of-Speech Error}

Part-of-speech errors arise when a lexical item is used in an inappropriate syntactic category.

\noindent \textbf{Example:}
\begin{itemize}[leftmargin=*]
    \item \textbf{Source:} \begin{CJK}{UTF8}{gbsn}他对这件事情的看法很不同意。\end{CJK}
    \item \textbf{Error:} \begin{CJK}{UTF8}{gbsn}不同意\end{CJK} (verb) is incorrectly used as an adjective
    \item \textbf{Corrected:} \begin{CJK}{UTF8}{gbsn}他对这件事情的看法很不同。\end{CJK}
\end{itemize}

\subsection{Word Order Error}

Word order errors involve incorrect constituent sequencing that violates Chinese syntactic constraints.

\noindent \textbf{Example:}
\begin{itemize}[leftmargin=*]
    \item \textbf{Source:} \begin{CJK}{UTF8}{gbsn}我把作业完成了已经。\end{CJK}
    \item \textbf{Error:} The aspectual adverb \begin{CJK}{UTF8}{gbsn}已经\end{CJK} is placed after the verb phrase
    \item \textbf{Corrected:} \begin{CJK}{UTF8}{gbsn}我已经把作业完成了。\end{CJK}
\end{itemize}

\subsection{Missing Component}

Missing component errors occur when obligatory syntactic elements are absent, resulting in incomplete or ill-formed sentences.

\noindent \textbf{Example:}
\begin{itemize}[leftmargin=*]
    \item \textbf{Source:} \begin{CJK}{UTF8}{gbsn}通过这次会议，使我们了解了情况。\end{CJK}
    \item \textbf{Error:} The sentence lacks a syntactic subject due to an improper causative construction
    \item \textbf{Corrected:} \begin{CJK}{UTF8}{gbsn}通过这次会议，我们了解了情况。\end{CJK}
\end{itemize}

\subsection{Redundant Component}

Redundancy errors involve semantically overlapping elements that should be eliminated.

\noindent \textbf{Example:}
\begin{itemize}[leftmargin=*]
    \item \textbf{Source:} \begin{CJK}{UTF8}{gbsn}他是一个很有天赋的有才华的人。\end{CJK}
    \item \textbf{Error:} \begin{CJK}{UTF8}{gbsn}有天赋的\end{CJK} and \begin{CJK}{UTF8}{gbsn}有才华的\end{CJK} are semantically redundant
    \item \textbf{Corrected:} \begin{CJK}{UTF8}{gbsn}他是一个很有天赋的人。\end{CJK}
\end{itemize}

\subsection{Connective Word Misuse}

Connective misuse errors involve violations of discourse-level conjunction conventions.

\noindent \textbf{Example:}
\begin{itemize}[leftmargin=*]
    \item \textbf{Source:} \begin{CJK}{UTF8}{gbsn}虽然天气很冷，但是他还是坚持锻炼。\end{CJK}
    \item \textbf{Error:} Redundant use of paired concessive connectives
    \item \textbf{Corrected:} \begin{CJK}{UTF8}{gbsn}虽然天气很冷，他还是坚持锻炼。\end{CJK}
\end{itemize}

\subsection{Ambiguous Reference}

Ambiguous reference errors occur when referential expressions lack clear antecedents.

\noindent \textbf{Example:}
\begin{itemize}[leftmargin=*]
    \item \textbf{Source:} \begin{CJK}{UTF8}{gbsn}小明和小红去看电影，他很喜欢这部电影。\end{CJK}
    \item \textbf{Error:} The pronoun \begin{CJK}{UTF8}{gbsn}他\end{CJK} has an unclear antecedent
    \item \textbf{Corrected:} \begin{CJK}{UTF8}{gbsn}小明和小红去看电影，小明很喜欢这部电影。\end{CJK}
\end{itemize}

\subsection{Semantic-Logical Inconsistency}

Semantic-logical errors involve violations of real-world knowledge or selectional restrictions.

\noindent \textbf{Example:}
\begin{itemize}[leftmargin=*]
    \item \textbf{Source:} \begin{CJK}{UTF8}{gbsn}这个盲人用眼睛仔细地观察着周围的环境。\end{CJK}
    \item \textbf{Error:} Logical contradiction between the subject and the action
    \item \textbf{Corrected:} \begin{CJK}{UTF8}{gbsn}这个盲人仔细地感受着周围的环境。\end{CJK}
\end{itemize}

\section{Faithfulness and Reliability of Distilled Chain-of-Thought}
\label{sec:cot_quality}

To explicitly address the reliability of the rationales distilled from the teacher model and to ensure they do not introduce hallucinatory explanations, we expanded our double-blind human evaluation post-metareview to systematically assess CoT faithfulness.

\subsection{Human Evaluation Setup and Agreement Metrics}
We randomly sampled 1,000 instances from our distilled CoT dataset. Three PhD annotators with expertise in Chinese linguistics independently evaluated the rationales. The evaluation focused on three fine-grained sub-dimensions: 
\begin{itemize}
    \item \textbf{Diagnostic\_Depth:} Whether the rationale accurately identifies the root cause of the grammatical or spelling error.
    \item \textbf{Information\_Integrity:} Whether the explanation is logically complete and avoids hallucinating non-existent linguistic rules.
    \item \textbf{Span\_Consistency:} Whether the localized error span strictly matches the proposed correction.
\end{itemize}

A rationale was considered ``Faithful'' (labeled as 1) if it successfully passed these criteria, and ``Unfaithful'' (labeled as 0) otherwise. The inter-annotator agreement metrics demonstrate high consistency among the experts, as shown in Table~\ref{tab:cot_agreement}.

\begin{table}[h]
\centering
\small
\begin{tabular}{lc}
\hline
\textbf{Metric} & \textbf{Score} \\ \hline
Semantic Agreement & 95.2\% \\
Average Pairwise $\kappa$ & 0.81 \\ \hline
\end{tabular}
\caption{Human evaluation metrics on 1,000 sampled CoT rationales. Semantic Agreement denotes the proportion of samples reaching a majority consensus ($\ge$ 2/3 labeled as Faithful). The Average Pairwise Cohen's $\kappa$ of 0.81 indicates almost perfect inter-annotator agreement.}
\label{tab:cot_agreement}
\end{table}

Of the 1,000 samples, 95.2\% reached a majority consensus as being faithful. To facilitate future research and transparency, we have open-sourced this annotated dataset along with the automated metric calculation scripts in our anonymous repository.

\subsection{Impact on Complex Edits}
Beyond human evaluation, we empirically verified the effectiveness of 
these faithful rationales on complex grammatical errors. We isolated a 
challenging subset of multi-span edits requiring substantial rewriting.

On this complex subset, the full CSRP model maintains a Recall of 31.2\%, significantly outperforming the SFT-only baseline (which lacks explicit CoT reasoning optimization), which degrades to a Recall of 24.5\%. This confirms that the explicit, faithful reasoning paths distilled during the CoT phase equip the model with the necessary diagnostic capability to handle genuinely complex, multi-span errors, rather than merely memorizing shallow surface mapping patterns.

\subsection{Summary}

These nine error categories cover the major grammatical and semantic phenomena in Chinese text correction. By explicitly modeling these error types through Chain-of-Thought rationales, our model learns not only to generate corrections, but also to provide linguistically grounded explanations for why a particular correction is warranted.

\end{document}